\documentclass[accepted]{uai2021} 

\usepackage[american]{babel}

\usepackage{natbib} 
\usepackage{amsfonts}       
\usepackage{nicefrac}     
\usepackage{microtype}      
\usepackage{subfigure}
\usepackage{amsmath}
\usepackage{amssymb}
\usepackage{adjustbox}
\usepackage{multirow}
\usepackage{algorithm}
\usepackage[noend]{algpseudocode}
\usepackage{accents}
\usepackage{epstopdf}
\usepackage{hyperref}
\graphicspath{{figures/}}

\newtheorem{theorem}{Theorem}

\newtheorem{definition}{Definition}

\DeclareMathOperator*{\argmax}{arg\,max}
\newcommand{\R}{{\mathbb R}}
\newcommand{\bbS}{{\mathbb S}}
\newcommand{\E}{{\mathbb E}}

\newcommand{\N}{{\mathcal N}}
\newcommand{\HH}{{\mathcal H}}
\newcommand{\A}{{\mathbb A}}

\DeclareMathOperator{\GP}{\mathcal{GP}}

\newcommand{\by}{{\mathbf{y}}}
\newcommand{\bx}{{\mathbf{x}}}
\newcommand{\bk}{{\mathbf{k}}}

\newcommand{\D}{{\mathcal{D}}}
\newcommand{\mI}{\mathsf{I}}
\newcommand{\mK}{\mathsf{K}}

\newcommand{\EIpu}{{\text{EIpu}}}
\newcommand{\EI}{{\text{EI}}}

\title{A Nonmyopic Approach to Cost-Constrained Bayesian Optimization}

\author[1]{Eric Hans Lee}
\author[2]{David Eriksson}
\author[3]{Valerio Perrone}
\author[3]{Matthias Seeger}
\affil[1]{SigOpt: An Intel Company; \texttt{eric.lee@intel.com}}
\affil[2]{Facebook; \texttt{deriksson@fb.com}}
\affil[3]{Amazon; \texttt{[vperrone, matthis]@amazon.de}}

\begin{document}
\maketitle

\begin{abstract}
Bayesian optimization (BO) is a popular method for optimizing expensive-to-evaluate black-box functions.
BO budgets are typically given in iterations, which implicitly assumes each evaluation has the same cost. 
In fact, in many BO applications, evaluation costs vary significantly in different regions of the search space.
In hyperparameter optimization, the time spent on neural network training increases with layer size; in clinical trials, the monetary cost of drug compounds vary; and in optimal control, control actions have differing complexities. 
\textit{Cost-constrained BO} measures convergence with alternative cost metrics such as time, money, or energy, for which the sample efficiency of standard BO methods is ill-suited. 
For cost-constrained BO, \textit{cost efficiency} is far more important than sample efficiency. 
In this paper, we formulate cost-constrained BO as a constrained Markov decision process (CMDP), and develop an efficient rollout approximation to the optimal CMDP policy that takes both the cost and future iterations into account. 
We validate our method on a collection of hyperparameter optimization problems as well as a sensor set selection application.
\end{abstract}

\section{INTRODUCTION}
\label{sec:introduction}
Bayesian optimization (BO) is a class of methods for global optimization of expensive black-box functions. 
In BO, a probabilistic surrogate model is used to approximate the objective and future evaluations are selected via an acquisition function. 
BO has been successfully applied to applications such as robotic gait control~\citep{calandra2016bayesian}, sensor set selection~\citep{garnett2010bayesian}, and neural network hyperparameter tuning~\citep{snoek2012practical,turner2021bayesian}.
BO is favored for these tasks because of its sample-efficient nature.
Achieving this sample-efficiency requires BO to balance exploration and exploitation. 
However, standard acquisition functions such as expected improvement (EI) are often too greedy in practice. 
As a result, they perform poorly on multimodal problems~\citep{hernandez2014predictive} and have provably sub-optimal performance in certain settings, e.g., bandit problems~\citep{srinivas2009gaussian}. 
A key research goal in BO is developing less greedy acquisition functions~\citep{shahriari2016botutorial}. 
Examples include predictive entropy search (PES)~\citep{hernandez2014predictive} or knowledge gradient (KG)~\citep{frazier2008knowledge}. 
\citet{lam2016bayesian} frame the exploration-exploitation trade-off as a balance between immediate and future rewards in a continuous state and action space Markov decision process (MDP). 
In this framework, \textit{non-myopic} acquisition functions are optimal MDP policies, and promise better performance by considering the impact of future evaluations up to a given BO budget (also referred to as the \textit{horizon}). 

While BO budgets are typically given in iterations, this implicitly measures convergence in terms of iteration count and assumes uniform evaluation cost. 
For many practical BO applications, evaluation costs may vary in different regions of the search space. 
For example, the time spent on neural network training increases with layer size; the cost of different drug compounds vary; and control actions in optimal control have differing complexities. 
In all these cases, standard BO is often unable to achieve fast convergence in terms of unit cost. 
Motivated by these examples, we develop methods that improve convergence when measured by cost. 
This cost may be time, energy, or money, and the goal is to minimize the objective given a cost budget. 

\begin{figure*}[!ht] 
    \centering
    \includegraphics[width=\textwidth]{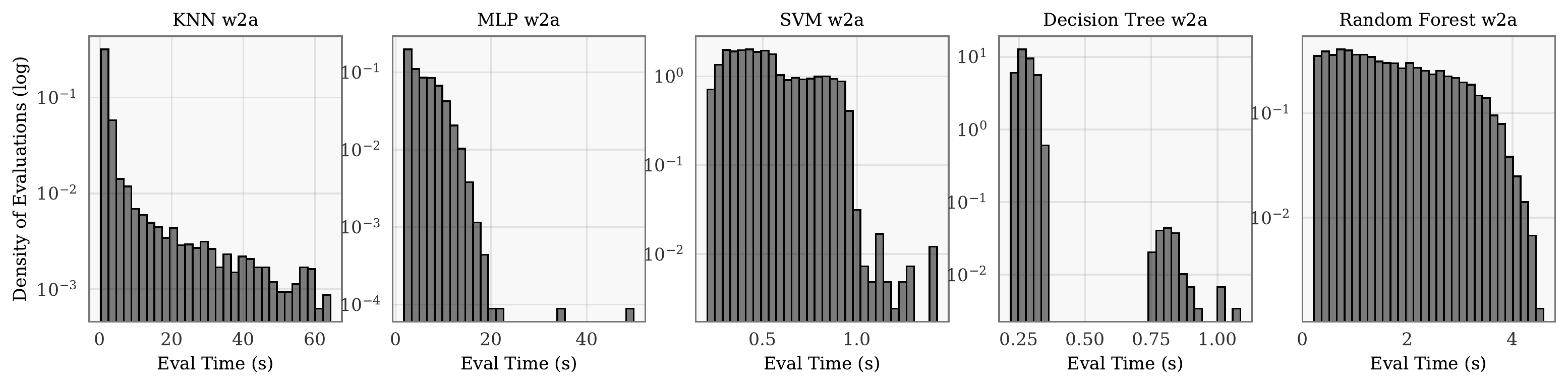}
    \caption{Runtime distribution, log-scaled, of $5000$ randomly selected points for the $k$-nearest-neighbors (KNN), Multi-layer Perceptron (MLP), Support Vector Machine (SVM), Decision Tree (DT), and Random Forest (RF) hyperparameter optimization problems, each trained on the OpenML w2a dataset ~\citep{OpenML2013}. The runtimes vary, often by an order of magnitude or more.}
    \label{fig:variations}
\end{figure*}

\textit{Cost-constrained BO} measures convergence with these alternative cost metrics for which standard BO methods are unsuited. 
We extend non-myopic BO to handle the cost-constrained setting. Our contributions follow:
\begin{itemize}
    \item We analyze failure modes of common approaches to cost-constrained BO, in which greedy behavior results in poor per-cost performance.  
    \item To avoid overly greedy behavior, we formulate cost-constrained BO as an instance of a constrained Markov decision process (CMDP). This formulation is a novel extension of recent research on non-myopic BO which uses a simpler Markov decision process (MDP) formulation. 
    \item We introduce an approximation to the optimal constrained MDP policy based on rollout of feasible trajectories. Rollout is a popular class of approximate MDP solutions in which future BO realizations and their corresponding values are simulated using the surrogate and then averaged.
    \item We validate the performance of our methods on a set of practical hyperparameter optimization problems and a sensor set selection problem.
\end{itemize}

\section{RELATED WORK}
\label{sec:related_work}
Most prior approaches to cost-constrained BO occur in the \textit{grey-box} setting, in which additional information about the objective is available. 
\textit{Multi-fidelity} BO is a widely studied setting in which fidelity parameters $s \in [0, 1]^m$, such as iteration count or grid size, are assumed to be a low-accuracy approximation of high-fidelity evaluations~\citep{forrester2007multi, kandasamy2017multi, poloczek2017multi, wu2019practical}. 
Increasing $s$ increases the accuracy at the expense of run time. 
In addition, Multi-fidelity methods are often application-specific. 
For example, Hyperband~\citep{li2017hyperband, falkner2018bohb, klein2016fast, klein2017fastbo} cheaply train many neural network configurations for a few epochs, and then prunes unpromising configurations. 

In \textit{multi-task} BO ~\citep{swersky2013multi}, hyperparameter optimization is first run on cheaper instances before considering more expensive ones. 
\citet{swersky2013multi} introduce a cost-constrained, multi-task variant of entropy search to speed-up optimization of logistic regression and latent Dirichlet allocation. 
Cost information is input as a set of cost preferences (e.g., a parameter $\bx_1$ is more expensive than a parameter $\bx_2$) by \citet{abdolshah2019costaware}, who develop a multi-objective, constrained BO method that evaluates cheap points before expensive ones, as determined by the cost preferences, to find feasible, low-cost solutions. 
These methods outperform their black-box counterparts by evaluating cheap proxies or cheap points before selecting expensive evaluations, which is accomplished by leveraging additional cost information inside the optimization routine. 

While all these methods demonstrate strong performance, they sacrifice generality and do not apply to black-box BO. 
Moreover, by relying on parallel resources, these techniques target time efficiency rather than compute time, cost, or energy efficiency. We also note that cost-efficiency has been modestly studied in the setting of active search, in which a active search is run with the constraint that the total cost of all queries must be less than $\tau$ \cite{jiang2019cost}.

Recent developments in nonmyopic BO account for the impact of future evaluations and are thus able to make better decisions~\citep{lam2016bayesian, lam2017lookahead, frazier2018, yue2019lookahead, lee2020efficient, jiang2020efficient}. These methods typically frame Bayesian optimization as a Markov decision process whose horizon is precisely the iteration budget. It is along these lines that we tackle the problem of nonmyopic, cost-constrained BO.   

\section{BACKGROUND AND MOTIVATION}
\label{sec:background}
\textbf{Gaussian process regression and BO:}
The goal in BO is to find a global minimizer of a continuous function $f(\bx)$ over a compact set $\Omega \subseteq \R^d$.
If $f(\bx)$ is expensive to evaluate, we want to rely on a sample-efficient optimization method.
BO uses a Gaussian process (GP) to model $f(\bx)$ from the data $\D_t = \{ (\bx_i,y_i) \}_{i=1}^t$.
We write this as $f(\bx) \sim \GP(\mu_t(\bx), \sigma^2_t(\bx))$, where $\mu_t(\bx), \sigma^2_t(\bx)$ are the GP mean and variance at $\bx$, respectively (see the supplementary materials for more details).
The next evaluation location $\bx_{t+1}$ is determined by maximizing an acquisition function $\Lambda(\bx \mid \D_t)$: $\bx_{t+1} =\argmax_{\Omega} \Lambda(\bx \mid \D_t)$.

\textbf{Cost-constrained BO:} BO's sample efficiency leads to fast convergence only if evaluations have similar costs, an assumption that is often not true in practice.
Cost-constrained BO is an important problem, and we argue that many BO problems in machine learning are, in fact, cost-constrained.
Figure~\ref{fig:variations} illustrates this by randomly evaluating $5000$ hyperparameter configurations for five common hyperparameter optimization problems (HPO).
The resulting evaluation times vary, sometimes by more than two orders of magnitude.
Moreover, the majority of evaluations are cheap, suggesting that significant cost savings may be achieved by \textit{using a cost-efficient instead of a sample-efficient optimizer}.

The de-facto cost-constrained method in the black-box setting is to normalize the acquisition by cost model $c(\bx)$.
This extends EI to \textit{EI per unit cost (EIpu)}:
\[
    \EIpu(\bx ) := \frac{\EI(\bx)}{c(\bx )},
\]
which is designed to balance the objective's cost and evaluation quality.
\citet{snoek2012practical} showed that EIpu can boost performance on a variety of HPO problems.

\begin{figure}[!ht]
    \centering
    \includegraphics[width=0.48\textwidth]{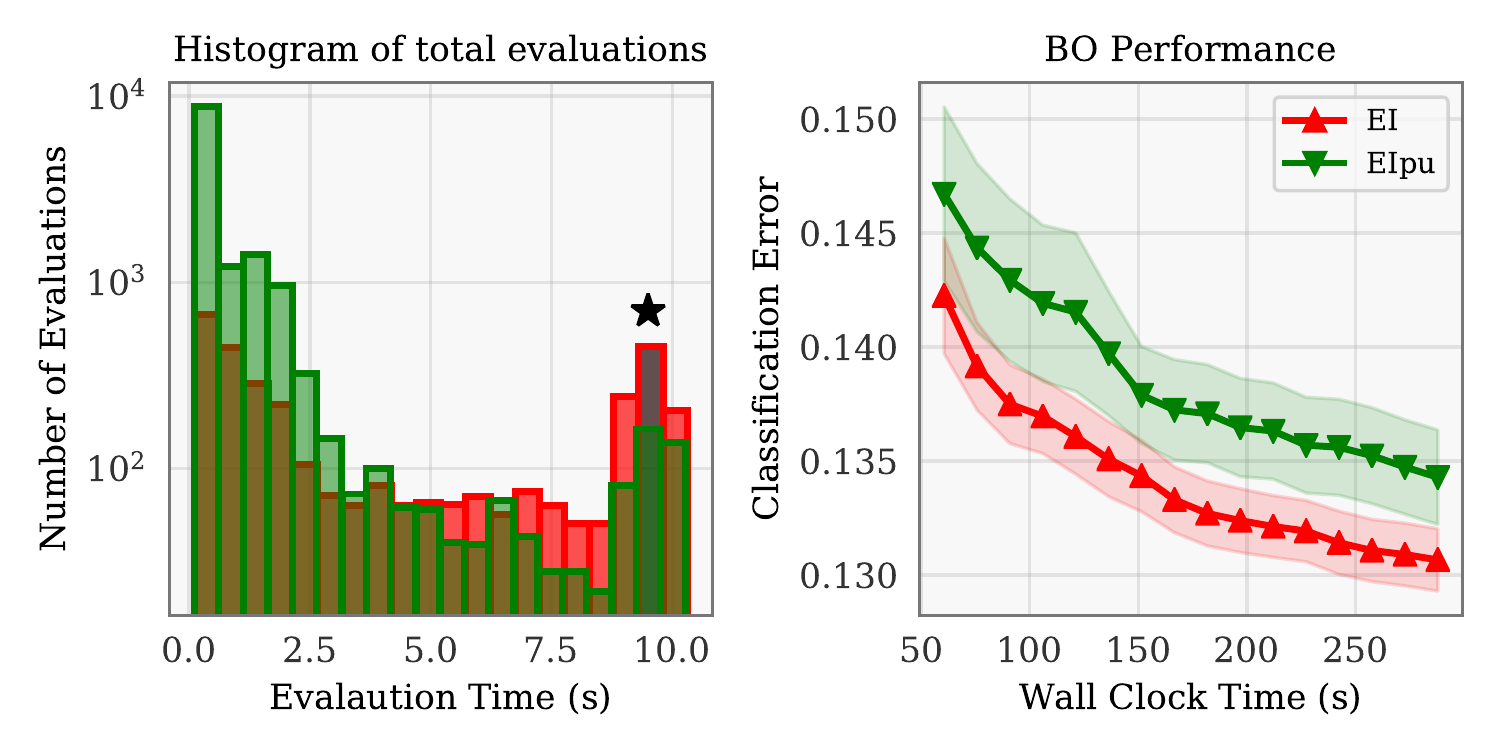}
    \caption{We run EI and EIpu on KNN. \textit{Left:} EIpu evaluates many more cheap points than EI, which evaluates more expensive points. The optimum's cost, one of the most expensive points, is a black star. \textit{Right:} EIpu performs poorly as a result. }
    \label{fig:knn_histogram}
\end{figure}

However, EIpu often demonstrates underwhelming performance.
We examine why this may occur in Figure~\ref{fig:knn_histogram}, in which EIpu (green) is slower than EI (red) at HPO of a $k$-nearest-neighbor model.
The empirical optimum, namely the best point over all trials (black star), has high cost ---thus, dividing by the cost penalizes EIpu \textit{away} from the optimum and diminishes its performance.
This is evident from the evaluation time histograms: EIpu evaluates many cheap points while EI evaluates fewer but more expensive points.
Due to its bias towards cheap points, EIpu is likely to only display strong results when optima are relatively cheap, which is problematic in the general black-box setting.
Indeed, one can adversarially increase the cost at the optimum to make EIpu perform poorly.

\textbf{Markov decision processes:} Nonmyopic BO frames the exploration-exploitation trade-off as a balance of immediate and future rewards in a finite horizon Markov decision process (MDP).
We use standard notation from~\citep{puterman2014mdp}: an MDP is the collection $<T, \bbS, \A, P, R>$.
Here, $T = \{0, 1, \ldots, h-1\}$, $h < \infty$ is the set of decision epochs, assumed finite for our problem.
The state space $\bbS$ encapsulates the information needed to model the system from time $t \in T$.
$\A$ is the action space.
Given a state $s \in \bbS$ and an action $a \in \A$, $P(s'| s, a)$ is the transition probability of the next state being $s'$.
$R(s, a, s')$ is the reward received for choosing action $a$ from state $s$, and ending in state $s'$.

A decision rule, $\pi_t: \bbS \rightarrow \A$, maps states to actions at time $t$.
A policy $\pi$ is a series of decision rules $\pi = (\pi_0, \pi_1, \ldots, \pi_{h-1})$, one at each decision epoch.
Given a policy $\pi$, a starting state $s_0$, and horizon $h$, we can define the expected total reward $V_h^\pi(s_0)$ as:
\[
    V_h^\pi(s_0) =  \E \left[ \sum_{t=0}^{h-1} R(s_t, \pi_t(s_t), s_{t+1})\right].
\]
In phrasing a sequence of decisions as an MDP, our goal is to find the optimal policy $\pi^*$ that maximizes the expected total reward, i.e., $\sup_{\pi \in \Pi} V_h^\pi (s_0)$, where $\Pi$ is the space of all admissible policies.

\begin{figure*}
    \centering
    \includegraphics[width=0.88\textwidth]{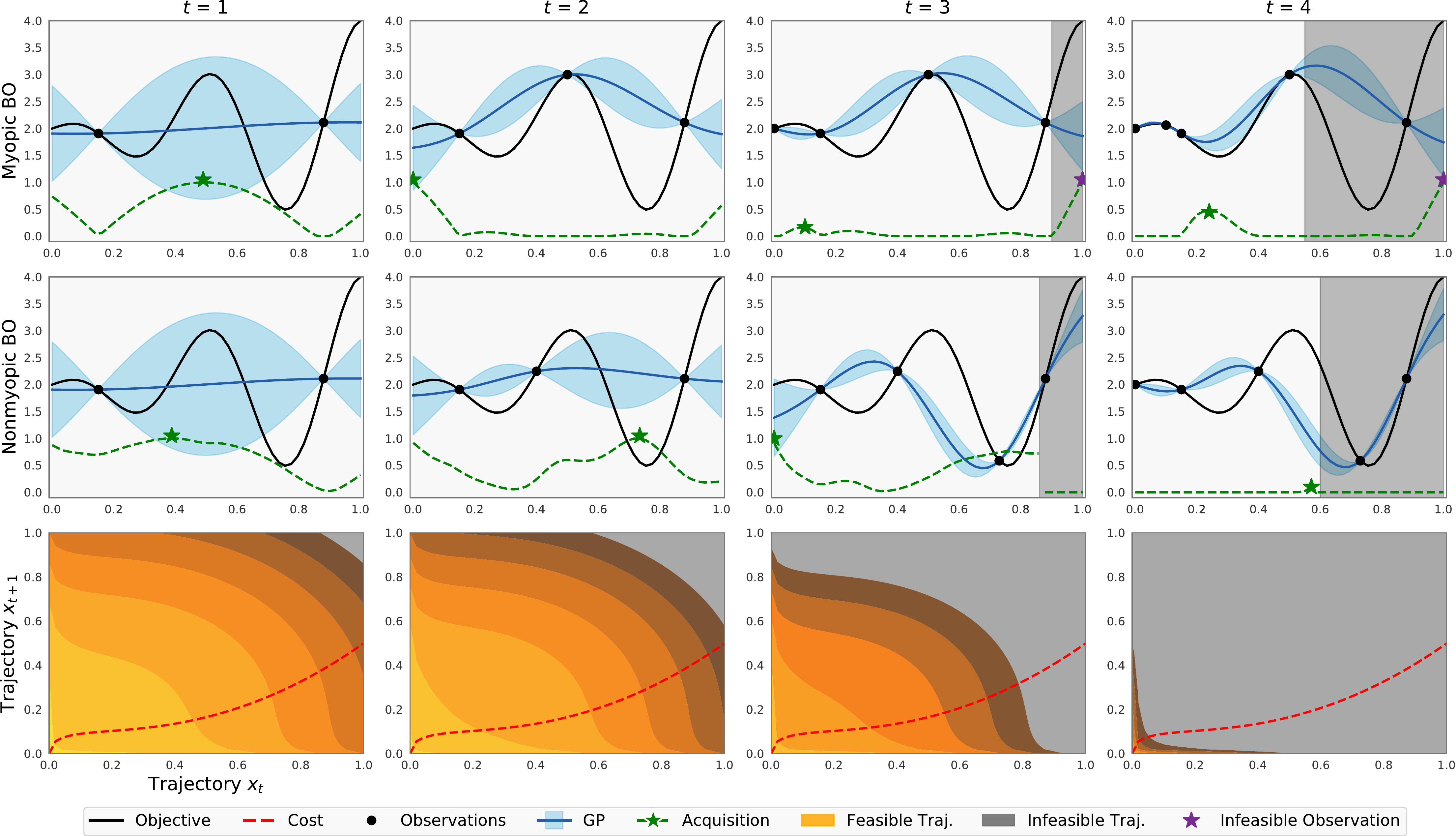}
    \caption{We illustrate the importance of nonmyopia on a carefully chosen toy problem. The top row depicts BO using a myopic acquisition; the middle row depicts nonmyopic, cost-constrained BO; the bottom row depicts feasible trajectories of length two. We run BO until the objective's minimum is infeasible. Myopic BO exhausts its budget before getting close to the global minimum. A nonmyopic approach accounts for the cost (and infeasibility) of future evaluations, and is able to sample the global minimum early. To save space, we only plot feasible trajectories for the myopic approach in the top row. The trajectory contour for nonmyopic BO is very similar.}
    \label{fig:bo_example}
\end{figure*}

\textbf{Constrained Markov decision processes:} A constrained Markov decision process (CMDP) is an MDP with an additional set of cost constraints~\citep{altman1999constrained, piunovskiy2006dynamic, bertsekas2005rollout}.
These costs, like MDP rewards, are accumulated through state by action until a certain horizon.
A CMDP extends an MDP, and is the collection $<T, \bbS, \A, P, R, C, \tau>$.
Here, $C(s, a, s') : \bbS \times \A \times  \bbS \rightarrow \R$ is a cost function measuring the cost of choosing action $a$ from state $s$, and ending in state $s'$.
$\tau$ is the cost constraint, and we assume without loss of generality that it is a positive scalar.

The cost function $C$ in a CMDP induces a cumulative cost function $C_h^\pi(s_0)$, which is analogous to a value function that replaces the reward with the cost:
\[
    C_h^\pi(s_0) =  \E \left[\sum_{t=0}^{h-1} C(s_t, \pi_t(s_t), s_{t+1}) \right].
\]
$C_h^\pi(s_0)$ measures total expected cost given a policy $\pi$, starting state $s_0$, and horizon $h$.
The goal in a CMDP is to find the optimal policy, defined as:
\begin{align*}
    &\pi^* = \argmax_{\pi} V_h^\pi(s_0), \\
    & \text{subject to } \; C_h^\pi(s_0) \leq \tau.
\end{align*}
In other words, we want to determine the policy that maximizes the expected reward subject to having cost less than $\tau$.
We refer readers to the standard CMDP treatment~\citep{altman1999constrained} for more information.
The notion of feasible trajectories is important when discussing both exact and approximate CMDP solutions.
A CMDP trajectory is a sequence of states and actions:
\[
    (s_0, a_0), (s_1, a_1), \dots, (s_{h-1}, a_{h-1}).
\]
A CMDP trajectory is said to be \textit{feasible} if it doesn't violate its cost constraint $\tau$ for any non-negative integer $0 \leq \ell \leq h-1$ less than horizon $h$:
\[
    \sum_{t=0}^{\ell} C(s_t, a_t, s_{t+1}) < \tau.
\]
For consistency, we can extend all feasible trajectories to have length $h$ by introducing an intermediate state and action which produce zero reward and cost (the formal equivalent of ``standing still'').
The set of all feasible trajectories is known as $G$.

\section{COST-CONSTRAINED BO}
\label{sec:cmdp}
We might think of cost-constrained BO as the following constrained optimization problem: 
\begin{equation*}
    \begin{split}
    & \min_{\textbf{X} \in 2^\Omega} \min_{\bx \in \textbf{X}} f(\bx) \\
    & \text{subject to } \sum_{\bx \in \textbf{X}} c(\bx) \leq \tau. 
    \end{split}
\end{equation*}
Because our optimization domain is $2^\Omega$, i.e., the power set of $\Omega$, we have a nested minimization problem. 
We assume $f(\bx)$ outputs not only its value, but also its evaluation cost determined by a cost function $c(\bx)$ ---which is also black-box. 
Our goal is to minimize $f(\bx)$ subject to the total evaluation cost not exceeding $\tau$. 
This is a pre-specified upper bound on the total cost, such as compute time, dollars, or energy consumption.\looseness-1

One recent research direction in BO has been the development of nonmyopic BO~\citep{lam2016bayesian, lam2017lookahead, yue2019lookahead, frazier2018, lee2020efficient, jiang2020efficient}, which account for the impact of future evaluations and are thus able to make better decisions. 
These decisions are computed by modeling BO as an MDP and then approximating its optimal policy. 
We aim to leverage this MDP framework to make similarly principled, nonmyopic decisions in the cost-constrained setting. 
We do this by extending the MDP to a CMDP, which takes into account variable evaluation costs. The next decision in this setting is a approximation to the optimal CMDP policy. 

Figure~\ref{fig:bo_example} illustrates the advantage of accounting for the cost of future evaluations with CMDP. 
The objective and cost, which have been carefully chosen, are plotted in black and red respectively, and the bottom row indicates feasible trajectories of length two as the optimization continues. 
The optimization domains are similarly shaded when evaluations become infeasible. 
A greedy approach\footnote{We show EI over EIpu for space's sake; the former performs better than the latter. The low cost on the left part of the domain causes EIpu to evaluate exclusively there.}, seen in the first row, does not account for the remaining budget, and is therefore unable to evaluate the global minimum before its becomes infeasible due to an insufficient budget. 
A nonmyopic policy, seen in the second row, better accounts for the cost of future evaluations; it sees that by the fourth iteration, the right side of the domain becomes infeasible, and decides to evaluate there earlier. As a result, it gets much closer to the global minimum. 

In the next section, we formalize our CMDP framework.
We note that our framework is vaguely related to BO with resources (BOR)~\citep{dolatnia2016bayesian}, who consider a partially observable MDP (POMDP) framework for BO when resource consumption of the objective varies, and when there might be multiple agents that can evaluate the acquisition function in parallel. 

\subsection{BO as a Constrained Markov Decision Process}
Given a deterministic cost function $c(\bx): \Omega \rightarrow \R^+$, a cost budget $\tau$, and a GP prior over the observation set $\D_t$ with mean $\mu_t$ and kernel $k_t$, we model $h$ steps of cost-constrained BO as the following CMDP: $<T, \bbS, \A, P, R, C, \tau>$. 

Here, $T$ is the set of decision epochs $\{0, 1, \ldots, h-1\}$ representing $h$ steps of BO. 
While we might want to use an infinite horizon, e.g., $h = \infty$ to iterate until our cost budget is exhausted, we assume a finite horizon for tractability. 
Our state space is the set of observations reachable from starting state $\D_t$ with $h$ BO steps, and the action space is $\Omega$; actions correspond to sampling a point in $\Omega$. 

The transition probabilities from state $\D_t$ to state $\D_{t+1}$, where $\D_{t+1} = \D_t \cup \{(\bx_{t+1}, y_{t+1})\}$, given an action $\bx_{t+1}$, are defined as:
\begin{align*}
    & P( \D_{t+1} \mid \D_t, \bx_{t+1}) \\
    & \sim \N(\mu^{(t)}(\bx_{t+1} ; \D_t), K^{(t)}(\bx_{t+1}, \bx_{t+1}; \D_t)). 
\end{align*}
In other words, the probability of transitioning from $\D_t$ to $\D_{t+1}$ is the probability of sampling $y_{t+1}$ from the posterior of $\GP(\mu_t, \sigma_t^2)$ at $\bx_{t+1}$.

Given an action and transition to a new state $\D_{t+1}$, our reward function is derived from the the EI criterion~\citep{jones1998ego}. 
Let $y_t^*$ be the minimum observed value in the observed set $\D_t$, i.e., $y_t^* = \min \{ y_0, \dots, y_t\}$. Then our reward is expressed as:
\begin{align*}
    & R(\D_t, \bx_{t+1}, \D_{t+1}) = ( y_t^* - y_{t+1})^+ \\
    & \equiv  \max (y_t^* - y_{t+1},0).
\end{align*}

Our CMDP cost is given by $c(\bx)$. 
We assume that this cost is deterministic and state-independent; it only depends on the action. 
In practice, the cost function may be learned as well. 
We emphasize that we assume a \textit{deterministic} cost function; the algorithms and theory we establish do not extend trivially to stochastic cost functions.
Finally, we assume a positive scalar constraint $\tau$. 
However, we can extend this to a vector-valued constraint. 
For example, in materials design, there might be a finite amount of each constituent component, each with its own budget~\citep{abdolshah2019costaware}.  

The expected total reward and cost of a policy $\pi$ are
\begin{align*}
    V_h^\pi(\D_k) &= \E \bigg[ \sum_{t=k}^{k + h-1} R(\D_t, \pi_t(\D_t), \D_{t+1}) \bigg] \\
    &= \E \bigg[ \sum_{t=k}^{k + h-1} (y^*_t - y_{t+1} )^+ \bigg] \nonumber, \\
    C_h^\pi(\D_k) &= \E \bigg[ \sum_{t=k}^{k + h-1} c(\pi_t(\D_t)) \bigg].
\end{align*}
More intuitively, $V_h^\pi(\D_k)$ is the expected reduction in the objective function using policy $\pi$, and $C_h^\pi(\D_k)$ is the accompanying expected cost. 
We can represent a trajectory though this CMDP as the sequence:
\[
    (\bx_k, y_k), (\bx_{k+1}, y_{k+1}), \dots, (\bx_{k+1}, y_{k+h}).
\]
As our cost is strictly positive, a trajectory $(\bx_k, y_k), (\bx_{k+1}, y_{k+1}), \dots, (\bx_{k+1}, y_{k+\ell})$ is feasible if $\sum_{i=k}^{k+\ell} c(\bx_i) \leq \tau$ for some $\ell \leq h$.

\section{METHODS}
\label{sec:methods}
CMDPs are considered far more difficult to solve than MDPs~\citep{altman1999constrained}, and the standard dynamic programming approach of~\citet{bertsekas1995dynamic} does not extend trivially ---Bellman's principle of optimality no longer applies.
Indeed, unlike the MDP case, the existence of an optimal policy is not guaranteed.
The standard CMDP solution is to solve a large linear program in the state and action spaces, but this is computationally intractable for all but the smallest problems.
The difficulty of solving CMDPs in the BO setting is made more difficult by the exponentially growing infinite state space, which consequently excludes standard solutions such as an exact solve on a discretized problem.

In this paper, we approximate the optimal CMDP policy through rollouts, which has been used successfully in the standard BO setting to improve performance over myopic acquisition functions~\citep{lam2016bayesian}.

\subsection{MDP Rollout}
Rollouts forward-simulate the value function of a fixed policy, and select the action yielding the maximal simulated reward.
We make this more precise as follows.
For a given current state $\D_k$, we denote our base rollout policy $\tilde \pi = (\tilde \pi_0, \tilde \pi_1, \ldots ,\tilde \pi_{h-1})$.
We introduce the notation $\D_{k,0}\equiv\D_k$ to define the initial state of our MDP and $\D_{k,t}$ for $1\leq t\leq h$ to denote the random variable that is the state at each decision epoch.
In the case of BO, each individual decision rule $\tilde{\pi}_t$ consists of maximizing the base acquisition function $B_t$ given the current state $s_t=\D_{k,t}$,
\[
    \tilde \pi_t = \argmax_{\bx \in \Omega} B_t(\bx \mid \D_{k,t}).
\]
Using this policy, we define the non-myopic acquisition function $\Lambda_h(\bx)$ as the rollout of $\tilde \pi$ to horizon $h$ i.e., the expected reward of $\tilde{\pi}$ starting with the action $\tilde{\pi}_0=\bx$:
\[
    \Lambda_{h}(\bx_{k+1}) := \E \bigg[ V^{\tilde \pi}_h (\D_k \cup \{(\bx_{k+1}, y_{k+1})\}) \bigg] ,
\]
where $y_{k+1}$ is the noisy observed value of $f$ at $\bx_{k+1}$.
$\Lambda_h$ performs better than the base policy in expectation for a correctly specified GP prior and for any base acquisition function.
This follows from standard results in the MDP literature \citep{bertsekas1995dynamic}.
If we can sample from the transition probability $P$, we can estimate the expected reward of $\tilde \pi$ through policy evaluation, i.e., Monte-Carlo integration:
\[
    V^{\tilde \pi}_h(s_0) \approx \frac{1}{N} \sum_{i=1}^N \bigg[\sum_{t=0}^{h-1} R(s^i_t, \tilde \pi_t(s^i_t), s^i_{t+1}) \bigg].
\]

\subsection{CMDP Rollout and Our Base Policy}
In the CMDP setting, rollout is a straightforward extension of rollout in the MDP setting. CMDP rollout also forward simulates of action and reward given a fixed base policy, except that it only keeps feasible trajectories in $G$ and discards infeasible trajectories~\citep{bertsekas2005rollout}.
In other words, this means that as we roll out a base policy $\tilde \pi$, we terminate either once we reach the horizon or violate the cost constraint.

There remains the question of what base policy to use; the performance of rollout depends on its base policy. We develop a base policy by considering the following two cases:

$\mathbf{h=1}$\textbf{:} Assume the argmax of EI has cost $c(\bx^*) \leq \tau$. The following policy $\pi$ is CMDP optimal:
    \[
    \pi(\D_t) = \bx^* = \argmax_{\bx \in \Omega} \EI(\bx \mid \D_t).
    \]
$\mathbf{h > 1}$\textbf{:} Assume the argmax of EI has cost $c(\bx^*) =\tau^*$ and there exists a point of small cost $c(\bx_{\epsilon}) = \epsilon$. If $ \tau = \tau^* + \epsilon$, then $\bx_{\epsilon}$ should be evaluated before $\bx^*$.
In the limit, a point that is free to evaluate should  be evaluated first.

A reasonable base policy should, at the minimum, satisfy these two cases.
For the first case, maximizing EI must necessarily be the last step in our base policy.
In the second case, we note that maximizing EIpu for the first rollout iteration will result in the desired behavior.
For simplicity's sake, we extend EIpu until the last iteration.
The base rollout policy $\tilde \pi = (\tilde \pi_0, \dots, \tilde \pi_{h-1})$ that we consider is therefore
\[
\tilde \pi_{t}(\D_t) =
    \begin{cases}
        \; \argmax_{\bx \in \Omega} \text{EIpu}(\bx \mid \D_{t}), & t < h-1,\\
        \; \argmax_{\bx \in \Omega} \text{EI}(\bx \mid \D_{t}), & t = h-1.
    \end{cases}
\]
In other words, $\tilde \pi$ rolls out $h-1$ steps of EIpu followed by a last step of EI.

This base policy has a few advantages. If $h=1$ and the budget is sufficient, it is CMDP optimal. If the cost is uniform, this is equivalent to rollout of EI, which has been shown to improve performance in standard BO~\citep{wu2019practical, lee2020efficient}.
Lastly, this base policy is consistent with an early exploration, late exploitation strategy, which is a common heuristic in multifidelity and multitask settings; EIpu tends to select cheaper points.
Therefore, $\tilde \pi$ starts by trying to select cheaper points and then ends with selecting a point that is likely more expensive.

\subsection{Theoretical Analysis}
If a base policy $\tilde \pi$ is \textit{sequentially consistent}, rollout in the MDP setting will perform better in expectation than the base policy itself ---this is known as the rollout improving property.
The same holds true in the CMDP setting if $c(\bx)$ is also deterministic. We define sequential consistency below.

\begin{definition} \label{def:sequentialconsistent}
    \citep{bertsekas1995dynamic}: A policy $\pi$ is sequentially consistent if, for every trajectory from any $s_0$:
    \[
        (s_0, a_0), (s_1, a_1), \dots, (s_{h-1}, a_{h-1}),
    \]
    $\pi$ generates the following trajectory starting at $s_1$:
    \[
        (s_1, a_1), (s_2, a_2) \dots, (s_{h-1}, a_{h-1}).
    \]
\end{definition}
Note that we have presented the simpler deterministic version of Definition~\ref{def:sequentialconsistent} for notational brevity; please refer to the appendix for the full stochastic version.

\begin{theorem}\label{theorem:constrained_mdp_rollout}
    \citep{bertsekas2005rollout}:
    In the CMDP setting, a rollout policy $\pi_{roll}$ does no worse than its base policy $\tilde \pi$ in expectation if $\tilde \pi$ is sequentially consistent i.e.,
    \[
        V^{\pi_{roll}}_h(s_0) \geq V^{\tilde \pi}_h(s_0).
    \]
    Thus, the value function of a rollout policy is always greater than the value function of the base policy.
\end{theorem}
To guarantee sequential consistency of our acquisition function, we need only consistently break ties if the acquisition function has multiple maxima.

\section{EXPERIMENTS}
\label{sec:experiments}
We compare CMDP rollout, which we compute via quasi-Monte Carlo integration, to EI and EIpu.
We use a GP with the Mat\'ern-$5/2$ ARD kernel to model both the objective and the cost function\footnote{We use a log-warped GP to model positive cost.}, and learn hyperparameters via maximum likelihood estimation.
When rolling out acquisition functions, we use L-BFGS-B using $5$ restarts, selected by evaluating the acquisition on a Latin hypercube of $10d$ points and picking the five best as starting points.
When comparing different replications we first need to interpolate the objective function values onto a set of discrete costs.
Given these interpolated value, we plot the mean with one standard deviation. Code to reproduce our experiments is found at \url{https://github.com/ericlee0803/lookahead_release}.

\subsection{Synthetic Problem}
\begin{figure}[!ht]
    \centering
    \includegraphics[width=0.48\textwidth]{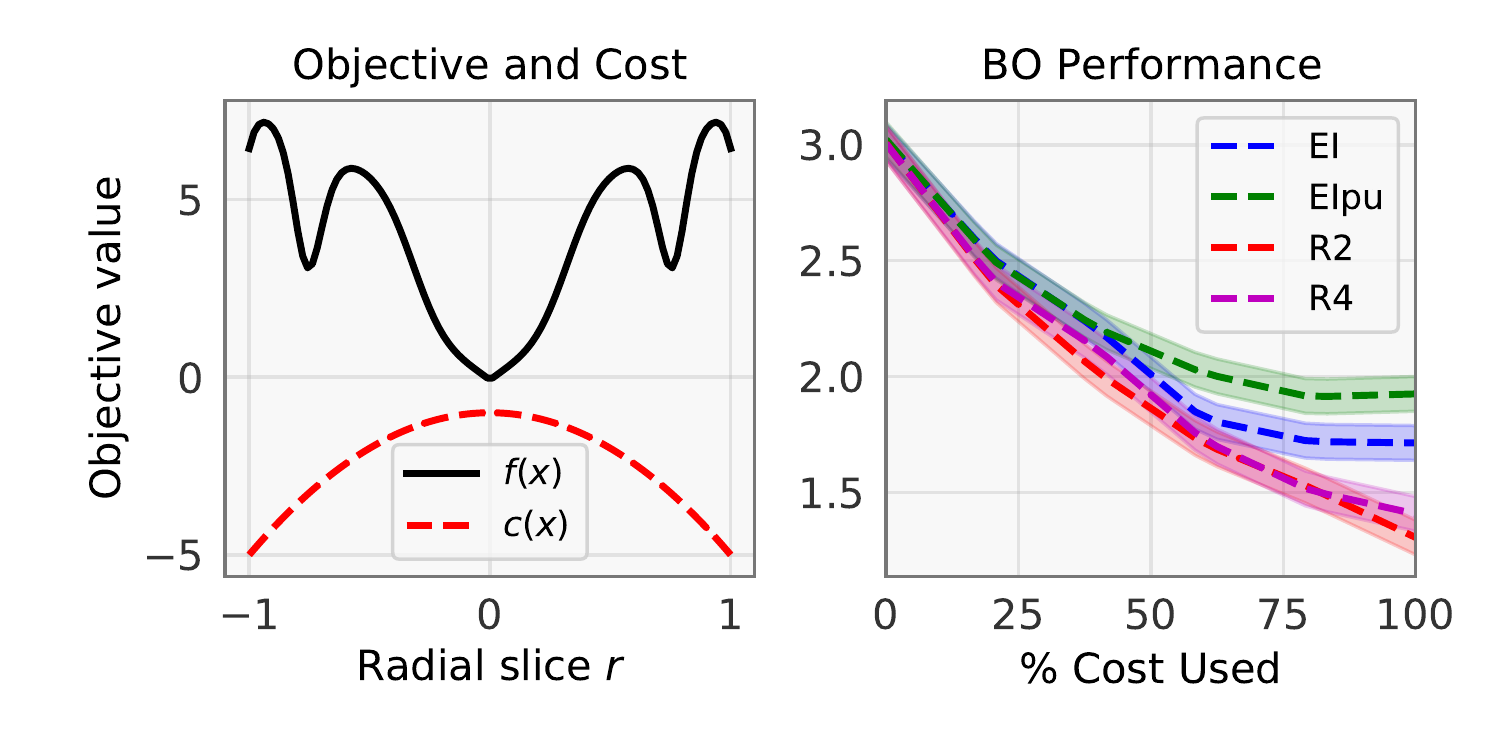}
    \caption{In this example, we examine a carefully chosen example showcasing the strength of the rollout approach. We consider the a multimodal objective whose most expensive point is the global minimum. EIpu performs worse than EI, and both tend to get stuck in cheaper, local minimum. Our rollout policy for horizons $2$ and $4$ performs better than both EI and EIpu}
    \label{fig:costrollout2d}
\end{figure}

In Figure~\ref{fig:costrollout2d}, we examine a carefully chosen synthetic example showcasing the strength of the rollout approach. We consider the cost-constrained optimization problem
\begin{align*}
    f(\bx) &= 10 \|\bx\|_2 \sin(2\pi \|\bx\|_2), \\
    c(\bx) &= 10 - 5\|\bx\|_2,
\end{align*}
in the domain $[-1, 1]^2$, and a budget of $150$. The cost function has been designed so that its maximum aligns with the minimum of the objective.
As we motivated earlier, EIpu struggles with these types of problems.
We run BO with EI, EIpu, and rollout with our base policy $50$ times and plot the results.
This is seen on the right, in which EIpu (green) performs worse than EI (blue).
However, rollout of our base policy, for horizons two and four in pink and red respectively, performs much better than both.

\subsection{Hyperparameter Optimization}
\begin{figure*}[!ht]
    \centering
    \includegraphics[width=\textwidth]{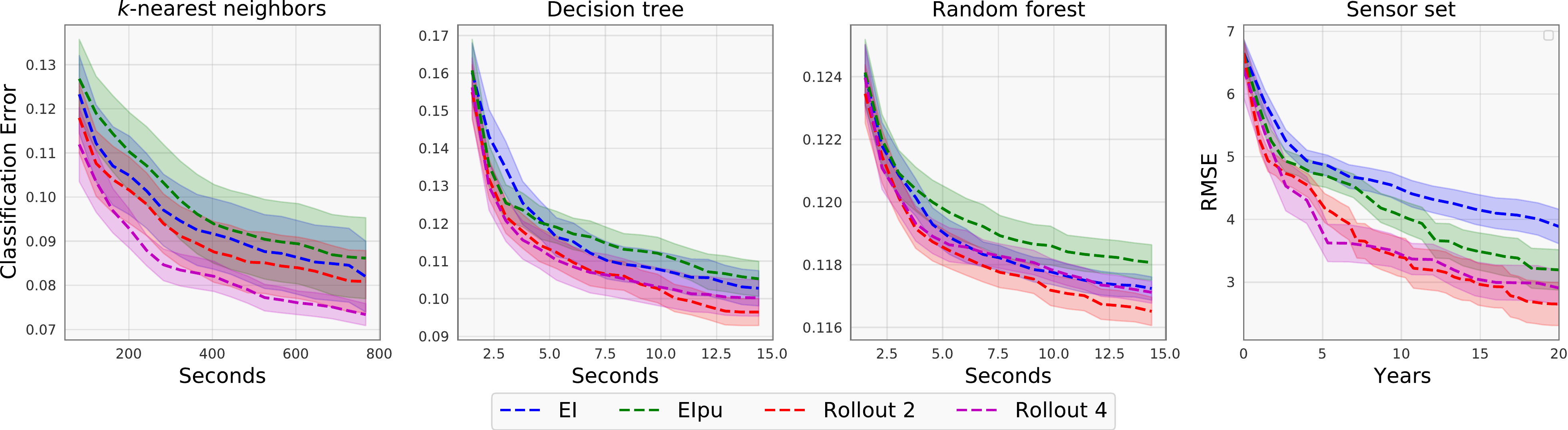}
    \caption{We compare the classification error among EI, EIpu, and our cost-constrained rollout for horizons $2$ and $4$. Rollout performs better than both EI and EIpu. Shaded areas represent one standard deviation around the mean. }
    \label{fig:hpo_results}
\end{figure*}
We compare rollout performance to EI and EIpu on HPO of three different models: $k$-nearest neighbors (KNN), decision trees, and random forests, with budgets of $800$, $15$, and $15$ seconds respectively. These are relatively small problems, chosen due to the number of replications required to show statistical significance.
All models are trained for $50$ replications on the OpenML w2a dataset~\citep{OpenML2013}. We compare the competing algorithms in terms of the best classification error achieved on the validation set.

\textbf{$k$-nearest neighbors:} The $k$-nearest neighbors algorithm is a class of methods used for classification of either spatially-orientated data or data with a known distance metric (i.e., data embedded in a Hilbert space).
We consider a $5d$ search space: dimensionality reduction percentage in $[1\mathrm{e}{-6}, 1.0]$ (log-scaled), type in \{Gaussian, Random\}, neighbor count in $\{1, 2, \ldots, 256\}$, weight function in \{Uniform, Distance\}, and distance in \{Minkowski, Cityblock, Cosine, Euclidean, L$1$, L$2$, Manhattan\}. We one-hot encode categorical variables.\looseness-1

\textbf{Decision Trees:} Decision trees are popular predictive models used in statistics, data mining, and machine learning.
In the case of classification, leaves represent class labels and paths represent sets of features that lead to those class labels.
During training, a tree is built by splitting the source set into subsets which constitute the successor children.
The splitting is based on a threshold that maximizes some notion of information gain such as entropy.
The depth of a decision tree is pre-specified.
We consider a $3d$ search space: tree depth in $\{1, 2, \ldots, 64\}$, tree split threshold in $[0.1, 1.0]$ log-scaled, and split feature size in $[1\mathrm{e}{-3}, 0.5]$ (log-scaled).

\textbf{Random Forests:} A random forest is a set $k$ of decision trees, and classifies based off the plurality decision generated from all its trees ---this technique is known as \textit{bagging}, and improves robustness in the classification algorithm.
We consider a $3d$ search space: number of trees in $\{1, 2, \ldots, 256\}$, tree depth in $\{1, 2, \ldots, 64\}$, and tree split threshold in $[0.1, 1.0]$ (log-scaled).

\subsection{Sensor Set Selection}

The sensor set selection problem~\citep{garnett2010bayesian} seeks to improve the predictive accuracy, as measured by the root mean squared error (RMSE), of a physical sensor network.
We denote a sensor network's configuration of $m$ sensors in $d$-dimensional space as $\mathbf{X}\in\R^{m \times d}$.

This configuration must be manually adjusted each time it is updated.
This is typically assumed to have uniform cost; we modify the problem to consider a \textit{sensor adjustment cost}.
We assume this cost is correlated with the distance each sensor in the network has to move; for simplicity, we assume the cost of any new sensor configuration $\mathbf{X}'$ is proportional to the straight-line distance $d(\mathbf{X}, \mathbf{X}') = \|\mathbf{X} - \mathbf{X}'\|_F$.
This is an example of a CMDP whose cost is not state-independent; our cost now depends on the prior state.

We consider a small sensor set selection problem using ten sensors.
Our objective is RMSE of the sensor predictions against ground truth weather data taken from UK Meteorological Office MIDAS surface stations~\citep{midas2012}.
The time budget is twenty years.

\subsection{Analysis}
We plot the resuts of our HPO and sensor set selection experiments in Figure~\ref{fig:hpo_results}, and find that CMDP rollout generally outperforms both EI and EIpu.
In this section we discuss key insights gained over the course of experimentation.

\textbf{Cost Modeling:} We found the cost function to be simpler to model than the objective function.
In practice, the cost may only depend on a few key parameters (e.g., tree depth).
Thus, using a vanilla GP to model the cost is inefficient ---a tailored (parametric) cost model or a GP that incorporates parameter importance into its lengthscale priors will likely lead to better results~\citep{lee2020cost, guinet2020}.

\textbf{Search Space Sensitivity:}
EIpu's performance depends on the correlation between objective value and cost.
Unsurprisingly, this correlation often depends on the search space in practice.
For example, assume a decision tree of depth $d$ achieves maximal classification error and that its training cost increases with depth.
(\textbf{i}) If the search space is $[1, d\,]$, the maximum will be the most expensive point and EIpu will perform poorly; (\textbf{ii}) if the search space is $[1, 2d\,]$, the maximum will be have middling cost and EIpu will perform moderately well; (\textbf{iii}) if the search space is $[1, 10d\,]$, the maximum will have cheap cost and EIpu will perform very well.
In our experiments, we found CMDP rollout to be more robust to the shape of the cost surface.
This is expected, as a CMDP optimal policy selects the point that maximally reduces the objective function given the cost constraint.

\section{CONCLUSION}
\label{sec:conclusion}
In this paper, we have shown the importance of cost-constrained BO and formulated it as an instance of a constrained Markov decision process (CMDP).
We developed a rollout algorithm using a cheap exploration, expensive exploitation base policy that performed better than EI and EIpu on three hyperparameter optimization problems and a sensor set selection problem.  

These investigations into cost-constrained BO are promising and we believe there are many interesting directions for future work. 
First, the overhead of the optimizer itself should be taken into account, especially in the context of HPO. 
While the overhead is negligible when the cost of evaluating the black-box is large (e.g., when training neural networks), future work could explore simpler heuristics to lower the overhead of using rollouts.
Second, we believe approximate solutions to CMDPs other than rollout are worth investigating. 
State aggregation and state truncation are classical methods in the MDP setting that reduce the state space according to the transition probabilities and~\citet{altman1999constrained} extends them to the CMDP setting. 
Consequently, we may approximate our model through state aggregation and state truncation and compute an exact solution via linear programming.\looseness-1

Finally, we have limited our discussion to the sequential BO setting. 
However, cost-constrained BO becomes significantly more complex in the batch setting, when evaluations are performed in parallel.
This is another interesting topic for future work.

\section*{Acknowledgements}
We want to thank Daniel Jiang for providing valuable feedback on the camera ready version of the paper.

\newpage
\bibliography{references}

\clearpage
\appendix
\label{sec:supplementary}
\appendix

\section{Gaussian process regression}
\label{sec:gps}
We place a GP prior on $f(\bx)$, denoted by $f \sim \GP(\mu, K)$, where $\mu:\Omega\to\R$ and $k:\Omega\times\Omega\to\R$ are the mean function and covariance kernel, respectively.
The kernel $k(\bx, \bx')$ correlates neighboring points, and may contain \textit{hyperparameters}, such as lengthscales that are learned to improve the quality of approximation~\citep{Rasmussen2006}. 
For a given $\D_t = \{ (\bx_i, y_i) \}_{i=1}^t$, we define:
\[
    \by = \begin{pmatrix}y_1 \\\vdots\\ y_t\end{pmatrix},
    \,\,
    \bk(\bx) = \begin{pmatrix} k(\bx, \bx_1) \\\vdots\\ k(\bx, \bx_t)\end{pmatrix},
    \,\,
    \mK = \begin{pmatrix}\bk(\bx_1)^\top \\\vdots\\ \bk(\bx_k)^\top\end{pmatrix}.
\]
We assume $y_i$ is observed with Gaussian white noise: $y_i = f(\bx_i) + \epsilon_i$, where $\epsilon_i \sim \N(0, \sigma^2)$.
Given a GP prior and data $\D_t$, the resulting posterior distribution for function values at a location $\bx$ is the Normal distribution $\N(\mu_t(\bx ; \D_t), \sigma^2_t(\bx; \D_t))$:
\begin{align*}
    \mu_t(\bx ; \D_t) &= \mu(\bx) + \bk(\bx)^\top(\mK + \sigma^2\mI_t)^{-1}(\by - \mu(\bx)), \\
    \sigma^2_t(\bx; \D_t) &= k(\bx, \bx) - \bk(\bx)^\top(\mK + \sigma^2\mI_t)^{-1}\bk(\bx),
\end{align*}
where $\mI_t \text{ is the $t \times t$ identity matrix}$. We use the Mat{\'e}rn 5/2 kernel in this paper:
\[
    k_{\text{5/2}}(\bx, \bx') = \alpha^2\left( 1 + \frac{\sqrt{5}}{\ell} + \frac{5}{3\ell^2}\right) \exp\left(-\frac{\sqrt{5}\|\bx-\bx'\|}{\ell} \right).
\]

\section{Theoretical Results}
If a base policy $\tilde \pi$ is \textit{sequentially consistent}, rollout in the MDP setting will perform better in expectation than the base policy itself
The same holds true in the CMDP setting if $c(\bx)$ is also deterministic. 

We first define a heuristic $\HH$ as a method to generate decision rules in epochs $\{0, \dots, h-1\}$, such that the resulting heuristic policy $\pi$ generated from $\HH$ on state $s$ is defined as
\[
\pi_{\HH(s)} = \{\pi_0 ^ {\pi_{\HH(s)}} \dots \pi_{h-1}^{\pi_{\HH(s)}}\}
\]

Having defined $\pi_{\HH(s)}$, we define sequential consistency for stochastic MDPs below.  

\begin{definition}
    \citep{goodson2017rollout}: A heuristic $\HH$ is sequentially consistent if, for every trajectory from any $s$ and all subsequent $s'$:
    \[
    \pi_{\HH(s)} = \pi_{\HH(s')},
    \]
    or in other words, that the decision rules generated from the heuristic are the same:
    \[
    \{\pi_0 ^ {\pi_{\HH(s)}} \dots \pi_{h-1}^{\pi_{\HH(s)}}\} = \ \{\pi_0 ^ {\pi_{\HH(s')}} \dots \pi_{h-1}^{\pi_{\HH(s')}}\}.
    \]
\end{definition}

This frames sequential consistency in terms of decision rules instead of as sample trajectories \cite{bertsekas1995dynamic}, which we did in the main text for notational clarity. However, the intuition of the theorem above is the same as its deterministic counterpart; a heuristic $\HH$ is sequentially consistent if it produces the same subsequent state $s'$ when started at any intermediate state of a path that it generates $s$.

\begin{theorem}
    \citep{bertsekas2005rollout}:
    In the CMDP setting, a rollout policy $\pi_{roll} = \pi^{\pi_{\HH(s)}}$ using sequentially consistent heuristic $\HH$ does no worse than its base policy $\tilde \pi$ in expectation.
    \[
        V^{\pi_{roll}}_h(s_0) \geq V^{\tilde \pi}_h(s_0).
    \]
    Thus, the value function of a rollout policy is always greater than the value function of the base policy. 
\end{theorem}
To guarantee sequential consistency of our acquisition function, we need only consistently break ties if the acquisition function has multiple maxima. 

\section{Additional Experiment}
 \begin{table}[h]
     \centering
     \begin{tabular}{| c||c|c|c|c |} 
          \hline & EI & EIpu & R2 & R4 \\ \hline
          mean & 0.140 & 0.139 & 0.137 & \textbf{0.131}\\
          std & 0.005 & 0.005 & 0.004 & \textbf{0.001}  \\ \hline
     \end{tabular}
     \caption{Rollout performance for horizons 2 and 4 outperformed both EI and EIpu. The best method is bolded.}
     \label{tab:supp_table}
 \end{table}

 Most of our HPO experiments were run on the OpenML w2a dataset. To sanity check our performance's robustness, we run the same HPO problem for k-nearest-neighbors with the OpenML a1a dataset. Rollout performance remained superior, and we record the mean and standard error in Table \ref{tab:supp_table}.

\end{document}